\documentclass[11pt,a4paper]{article}
\usepackage[hyperref]{emnlp2021}
\usepackage{booktabs}
\usepackage{mathrsfs}
\usepackage{makecell}
\usepackage{multirow}
\usepackage{times}
\usepackage{latexsym}
\usepackage{color}
\usepackage{bm}
\usepackage{latexsym}
\usepackage{amssymb,amsmath}
\usepackage{mathrsfs}
\usepackage{graphicx}
\usepackage{caption}
\usepackage{url}
\usepackage{url}
\usepackage{multirow}
\usepackage{verbatim}
\usepackage{amsmath}
\usepackage{subcaption}
\usepackage{array}
\usepackage{amsfonts}
\usepackage{algorithm}
\usepackage[noend]{algorithmic}
\usepackage{graphicx}
\usepackage{listings}
\usepackage{verbatim}
\usepackage{varwidth}
\usepackage{xcolor}
\usepackage{commath}
\usepackage[textsize=tiny]{todonotes}
\usepackage[normalem]{ulem}

\usepackage{xcolor}
\usepackage{pgfplots}
\usepackage{tikz}
\usepackage{filecontents}
\usepackage{pgfplotstable}
\usepackage{tikz-qtree}
\usepackage{enumitem}

\usepackage{microtype}

\newcommand{\wei}[1]{\textcolor{black}{#1}}

\title{Global Explainability of BERT-Based Evaluation Metrics by\\ Disentangling along Linguistic Factors}

\author{Marvin Kaster, Wei Zhao, Steffen Eger\\
    Natural Language Learning Group (NLLG) \\
    Technische Universit\"at Darmstadt, Germany \\
    {\tt marvin.kaster@stud.tu-darmstadt.de} \\
    {\tt \{zhao,eger\}@aiphes.tu-darmstadt.de}\\
  }
  
\date{}

\begin{document}
\maketitle
\begin{abstract}
Evaluation metrics are a key ingredient for progress of text generation systems. In recent years, several BERT-based evaluation metrics have been proposed (including BERTScore, MoverScore, BLEURT, etc.) which correlate much better with human assessment of text generation quality than BLEU or ROUGE, invented two decades ago. However, little is known what these metrics, which are based on black-box language model representations, actually capture (it is typically assumed they model semantic similarity). In this work, we use a simple regression based global explainability technique to disentangle metric scores along linguistic factors, including semantics, syntax, morphology, and lexical overlap. We show that the different metrics capture all aspects to some degree, but that they are all substantially sensitive to lexical overlap, just like BLEU and ROUGE. This exposes limitations of these novelly proposed metrics, which we also highlight in an adversarial test scenario. 
\end{abstract}

\section{Introduction}
Evaluation metrics are a key ingredient in assessing the quality of text generation systems, be it machine translation, summarization, or conversational AI models. Traditional evaluation metrics in machine translation and summarization, BLEU and ROUGE \citep{papineni-etal-2002-bleu,lin-2004-rouge}, have measured lexical $n$-gram overlap between system prediction and a human reference. While simple and easy to understand, early on, 
limitations of such lexical overlap metrics have been recognized 
\citep{callison-burch-etal-2006-evaluating}, e.g., in that they 
can only measure surface level similarity, and they are especially inadequate when it comes to assessing current high-quality text generation systems \cite{rei-etal-2020-comet,mathur-etal-2020-tangled,marie-etal-2021-scientific}. 

Recently, a class of novel evaluation metrics based on BERT and 
its variants 
has been explored that correlates much better with human assessments of translation quality. For example, BERTScore \citep{zhang2019bertscore}, MoverScore \citep{zhao2019moverscore}, BLEURT \citep{sellam2020bleurt}, XMoverScore \citep{zhao-etal-2020-limitations}, and COMET \citep{rei-etal-2020-comet} all use large-scale pretrained language models, but differ in whether they compare hypotheses to  references, 
to source texts, 
to both, 
on the one hand, and whether they use human scores for supervision 
or not, 
on the other. 
Since these models 
all leverage 
large-scale language models 
which have pushed the state-of-the-art in many areas of NLP, 
their success comes with little surprise. 

To better understand these novel metrics based on black-box language 
representations is a prerequisite for identifying their limitations, e.g., to adversarial inputs. For example, if an evaluation metric is sensitive to lexical overlap, it can be fooled by using the same words but in different order.

In this work, we fill the existing `explainability gap' and introspect 
linguistic properties encoded in BERT-based evaluation metrics. 
Although there is already considerable work on introspecting and understanding BERT (see \citet{rogers-etal-2020-primer} for an overview), e.g., via probing  \cite{tenney-etal-2019-bert}, 
analyzes by~\citet{hewitt-liang-2019-designing,eger-etal-2020-probe, ravichander-etal-2021-probing} indicate that probing results (based on supervision) are not always trustworthy. 
More importantly, the modern evaluation metrics sketched above rely on at least two factors: BERT (or its variants) and different aggregation schemes, such as Earth Mover Distance \citep{kusner2015word,zhao2019moverscore} or greedy alignment \citep{zhang2019bertscore}, on top of BERT. Understanding BERT alone is thus not sufficient for explaining BERT-based evaluation metrics.

Here, we present a simple global explanation technique of BERT-based evaluation metrics which disentangles them on prominent linguistic factors, \emph{viz.}, syntax, semantics, morphology and lexical overlap. We find that all metrics capture these linguistic aspects to certain (but differing) degrees, and they are particularly sensitive to lexical overlap, which makes them prone to similar adversarial fooling \citep[cf.][]{li-etal-2020-bert-attack,eger-etal-2019-text,keller-etal-2021-bert} as BLEU-based lexical overlap metrics. 
Overall, our contributions are: 
\begin{itemize}[topsep=5pt,itemsep=0pt,leftmargin=*]
    \item We disentangle a multitude of current BERT-based evaluation metrics on four linguistic factors using linear regression.  
    \item We show that all metrics are sensitive to all factors and especially to lexical overlap, as confirmed both by the linear regression and an adversarial  experiment. 
    \item Based on the insight that different metrics capture different linguistic factors to varying degrees, we ensemble metrics and identify an average improvement of between 8 and 13\% for the most heterogeneous metrics. 
\end{itemize}
\section{Related work}

Our work concerns reference-based and reference-free metrics, on the one hand, and model introspection (or `explainability'), on the other.

\paragraph{Evaluation metrics for Natural Language Generation}
In the last few years, several strong performing evaluation metrics have been proposed, the majority of which is based on BERT and similar high-quality text representations. They can be differentiated 
along two dimensions: (i) the input arguments they take, and (ii) whether they are supervised or unsupervised. \emph{Reference-based} metrics compare human references to system predictions. Popular metrics are BERTScore \citep{zhang2019bertscore}, MoverScore \citep{zhao2019moverscore}, and BLEURT \citep{sellam2020bleurt}. In contrast, \emph{reference-free} metrics directly compare source texts to system predictions, thus they are more resource-lean. 
Popular examples are XMoverScore \citep{zhao-etal-2020-limitations}, Yisi-2 \citep{lo-2019-yisi}, KoBE \citep{gekhman-etal-2020-kobe}, and SentSim \citep{song-etal-2021-sentsim}. \citet{rei-etal-2020-comet} use all three information signals: source text, hypotheses and human references. 
There are also reference-free metrics outside the field of machine translation; for example, SUPERT \citep{gao-etal-2020-supert} for summarization. 
\emph{Supervised} metrics train on human sentence-level scores, e.g., Direct Assessment (DA) scores 
or post-editing effort (HTER) for MT. These metrics include BLEURT and COMET \citep{rei-etal-2020-comet}. In MT, most metrics from the so-called `Quality Estimation' (QE) tasks are also supervised reference-free metrics, e.g., TransQuest \citep{ranasinghe-etal-2020-transquest} and BERGAMOT-LATTE \citep{fomicheva-etal-2020-bergamot}. 
\emph{Unsupervised} metrics require no such supervisory signal (e.g., MoverScore, BERTScore, XMoverScore, SentSim).  
 
\paragraph{Model introspection} There has been a recent surge in interest in explaining deep learning models. 
The techniques for explainability differ in whether they 
provide justification or information for 
model outputs on individual instances (\textit{local explainability}) or focus on a model as a whole and disclose its internal structure (\textit{global explainability}) 
\cite{danilevsky-etal-2020-survey}. 
Popular examples 
for local explainability 
are LIME \cite{lime} and SHAP \cite{NIPS2017_7062} that find  
features from the input (such as particular words) relevant to model outputs. 

Concerning (global) interpretability of text representations, 
previous works~\cite{AdiKBLG17, conneau-etal-2018-cram} introspect the properties encoded in vector representations through probing classifiers---trained  
on external data to perform a certain linguistic task, such as inducing the dependency tree depth from a text representation (of a sentence).
\citet{tenney-etal-2019-bert} extend this idea by inspecting BERT representations layer-by-layer, and find that BERT captures more semantic information in its higher layers and more syntactic and morphological information in its lower layers. However, probing results are not always trustworthy due to the sensitivity to probing design choices, e.g., 
data size and classifier choices~\cite{eger-etal-2020-probe},  and data artefacts~\cite{ravichander-etal-2021-probing}. More importantly, evaluation metrics use BERT differently: 
some are supervised and others are unsupervised, some fine-tune BERT on semantic similarity datasets, and they generally differ on how they aggregate and compare BERT representations. 
This means to understand these metrics 
it does not suffice to understand BERT alone.

In our work,  
we disentangle BERT based evaluation metrics along linguistic factors as a form of global explainability of those metrics. This yields insights into which linguistic information signals specific metrics use, in general, and may 
expose their limitations.

\section{Our approach}
\label{sec:approach}

In our scenario, we consider different metrics $\mathfrak{m}$ taking two arguments and assigning them a real-valued score
\begin{align*}
    \mathfrak{m}: (x,y) \mapsto s_{\mathfrak{m}} \in \mathbb{R}
\end{align*}
where $x$ and $y$ are source text and hypothesis text, respectively (for so-called \emph{reference-free} metrics) or alternatively $x$ and $y$ are reference and hypothesis text, respectively (for so-called \emph{reference-based} metrics). The scores $s$ that metrics assign to $x$ and $y$ can be considered the  \emph{similarity} between $x$ and $y$ or \emph{adequacy} of $y$ given $x$. In our experiments in Section \ref{sec:experiments}, we will focus on machine translation (MT) as use case; it is arguably the most popular and prominent text generation task. Thus, $x$ is either a (sentence-level) source text in one language and $y$ the corresponding MT output, or $x$ is the human reference for the original source text.  

To better understand evaluation metrics, we 
decompose their scores $s_{\mathfrak{m}}$ along 
multiple linguistic factors. 
An example is outlined in Table \ref{table:example}. 

\begin{table*}[!htb]
\small 
\centering
\begin{tabular}{p{4.8cm}|p{4.8cm}|cccc}
     \toprule
     \textbf{Hypothesis} ($y$) & \textbf{Reference}/\textbf{Source} ($x$) & \textbf{SEM} & \textbf{SYN} & \textbf{LEX} & \textbf{MOR}\\ \midrule
     It is a boy , likes to sport , but it cannot do it because of their very. & He is a boy, he likes sports but he can't take part because of his knee. &  -1.57  & 0.98 & -0.59 & -0.87 \\ \midrule
     
     Zwei Besatzungsmitglieder galten als vermisst.	 & Two crew members were regarded as missing.  &	0.83 & 0.99  & 0.46 & -2.40\\
     \bottomrule
\end{tabular}
\caption{Example setups with normalized semantic, syntactic, lexical overlap and morphological scores.}
\label{table:example}
\end{table*}

We follow a long line of research in the applied sciences, and use a \emph{linear model} to explain a \emph{target variable} (in our case, the metric score), also called \emph{response variable}, in terms of multiple \emph{regressors}, also called \emph{explanatory variables}. That is, we estimate the linear regression  
\begin{equation}\label{eq:1}
\begin{split}
\mathfrak{m}(x,y) =  \alpha \cdot\textit{sem}(x,y) + \beta \cdot\textit{syn}(x,y) \\   + \,\gamma \cdot\textit{lex}(x,y) + \delta\cdot \textit{morph}(x,y)  +\,\epsilon
\end{split}
\end{equation}

Here, $\text{sem}(x,y)$, $\text{syn}(x,y)$,  $\text{morph}(x,y)$ and $\text{lex}(x,y)$  are scores which describe 
the semantic, syntactic, morphological, and lexical similarity
between the two argument sentences. The real coefficients $\alpha$, $\beta$, $\gamma$ and $\delta$ 
are the regressors' weights, estimated from data. Finally, $\epsilon$ is an error term. 

Linear regression assumes a linear relationship between the target variable and the regressors. It may fail when the relationship is non-linear, but its simple model structure provides interpretability: a larger positive coefficient means the respective regressor has higher positive impact on the target variable (fixing all other variables), a coefficient close to zero means no linear relationship, and a larger negative coefficient means an inverse linear relationship between regressor and target variable. 

The coefficient of determination $R^2$ describes how well the regression model reflects the data. It is defined as 
\[
R^2 = 1 - \frac{\text{SSE}}{\text{SST}}
\]
where \textit{SSE} denotes the sum of squared errors 
and \textit{SST} denotes the sum of squared totals. 
They are defined as
\begin{align*}
\text{SSE} = \sum_i (y_i - \hat{y_i})^2,\quad 
\text{SST} = \sum_i (y_i - \overline{y}_i)^2
\end{align*}
respectively, where $\hat{y_i}$ is the prediction of the model, $y_i$ is the true score, and $\overline{y}_i$ is the mean,  $\overline{y}_i = \frac{1}{N} \sum^N_i y_i$. 
$R^2$ is 1 for a perfect fit, 0 if it always predicts the mean and negative if the model is worse than this baseline.

To ensure comparability of the different regressions, 
we normalize the scores of our regressors and the response variable 
with the z-normalization, i.e., subtracting the mean and dividing by the standard deviation, per variable. 
In the following, we define our regressors. 

\paragraph{Semantic score (SEM)} 
The semantic scores are provided by the datasets and were annotated by humans who rated e.g.\ the translation quality. See Section \ref{sec:datasets} for details.

\paragraph{Syntactic score (SYN)} 
\begin{figure*}
    \centering
    \scalebox{.92}{
    
\begin{subfigure}{.20\textwidth}
\centering
\Tree[ .knew He [ .little very ] [ .him about ] ]   
\end{subfigure}%
\begin{subfigure}{.20\textwidth}
\centering
\Tree[ .X X [ .X \textcolor{red}{X} ] [ .X \textcolor{blue}{X} ] ]   
\end{subfigure}%
\begin{subfigure}{.20\textwidth}
\centering
\Tree[ .X X X  [ .X \textcolor{blue}{X} ] ]      
\end{subfigure}%
\begin{subfigure}{.20\textwidth}
\centering
\Tree[ .X X X X ]     
\end{subfigure}%
\begin{subfigure}{.20\textwidth}
\centering
 \Tree[ .unknowable he was very ]
\end{subfigure}%
}
\caption{
Monolingual tree edit distance example.
The left-most tree is the first sentence of the sentence pair and the  right-most tree is the second. 
To transform the left-most sentence into the right-most sentence, two leaves are removed. The unnormalized tree edit distance is thus 2. The normalized score is $1-\frac{2}{6+4} = 0.8$. 
}
\label{fig:ted}
\end{figure*}
\begin{figure*}
    \centering
    \scalebox{0.92}{
\begin{subfigure}{.20\textwidth}
\centering
\Tree[ .mother it is a great ]   
\end{subfigure}%
\begin{subfigure}{.20\textwidth}
\centering
\Tree[ .X X \textcolor{blue}{X} \textcolor{red}{X} X  ]      
\end{subfigure}%
\begin{subfigure}{.20\textwidth}
\centering
\Tree[ .X X \textcolor{blue}{X} [ .X \textcolor{red}{X} ] ]      
\end{subfigure}%
\begin{subfigure}{.20\textwidth}
\centering
 \Tree[ .X X [ .X \textcolor{blue}{X} \textcolor{red}{X} ] ]
\end{subfigure}%
\begin{subfigure}{.20\textwidth}
\centering
 \Tree[ .ist Sie [ .Mutter eine großartige ] ]
\end{subfigure}%
}

\caption{Cross-lingual tree edit distance example. 
To transform the left-most into the right-most sentence, two leaves (shown in red and blue) are moved to other locations.
This takes 2 operations. 
The normalized score is $1-\frac{2}{5+5} = 0.8$.
}
\label{fig:ted_de}
\end{figure*}
To 
measure the syntactic similarity of the argument sentences $x$ and $y$, 
we compare their dependency trees. 
Both sentences are parsed by the Stanford dependency parser \citep{chen2014fast}. Then, the tree edit distance (TED) \citep{bille2005survey} between the resulting trees is calculated. As an extension of string edit distance, TED measures how many operations are necessary to transform one tree into the other. Only the structures of the trees are considered in the calculation. 
We ignore the actual words. 

We normalized the TED to 
ensure comparability between sentences with different lengths \citep{zhang1989simple}. The final score is calculated as 
\[
\textit{syn}(x,y) = 1-\frac{\textit{TED}}{l_x+l_y}
\]
where 
$l_x$ and $l_y$ are the lengths of the sentences.
Figure \ref{fig:ted} shows an example of the TED calculation for sentences in the same language (monolingual) and Figure \ref{fig:ted_de} shows a cross-lingual example. 

\paragraph{Lexical overlap score (LEX)} 
We measure the lexical overlap between $x$ and $y$ 
by the BLEU score \citep{papineni2002bleu}: $\text{BLEU}_n$ calculates the precision based on how many $n$-grams of one sentence can be found in the other sentence. In the experiments below, we use $\text{BLEU}_1$. Using unigrams assures that word order is ignored. The simple precision count is modified so that identical words are only counted once.

For monolingual reference-based metrics, the BLEU score is calculated directly on the sentence pairs. To use BLEU 
for cross-lingual reference-free metrics,  
we \wei{choose to} translate the non-English sentences 
into English via Google Translate, \wei{as it remains unclear how else to define lexical overlap between  
sentences from different languages.} 
We compute BLEU scores on original English and translated sentence pairs.

\begin{table}[!htb]
    \centering
    { 
    \small 
    \begin{tabular}{cc|c}
    \toprule
        \textbf{Word 1} & \textbf{Word 2} & \textbf{UD Tags}\\
        \midrule
         reached  & combined  & {\small Tense=Past VerbForm=Part}\\
        stay &  sein & {\small VerbForm=Inf}\\
         \bottomrule
    \end{tabular}
    \caption{\wei{Example of a morphological lexicon that contains word pairs with identical morphological UD tags, on which we finetune FastText word embeddings.}}
    \label{tab:morph}
    }
\end{table}

\paragraph{Morphological score (MOR)} 
We introduce a morphological score $\textit{morph}(x,y)$. To do so, we use static FastText word embeddings \citep{bojanowski2017enriching} and 
increase  
morphological information in the original embeddings:  
(a) first, we produce two morphological lexicons based on words from WMT and STS, each containing word pairs with identical UD morphological tags \citep{nivre2020universal} (see Table \ref{tab:morph}). (b) Then, we finetune/retrofit the embeddings on the morphological lexicons using the method described in \citet{faruqui:2014:NIPS-DLRLW}, 
so that words 
with the same morphological tags have more similar representations.

The final morphological score for a sentence pair is the cosine similarity between two averaged sentence embeddings over refined word vectors of each sentence.
Note that, while we refer to these embeddings as morphological, they actually capture multiple linguistic factors and can only be considered \emph{more} morphological than standard static vector spaces. 

If the overlap of morphological features between a language pair is very low, the morphological score will not be meaningful. We exclude the morphological score   
for such language pairs. 
\section{Experiments}\label{sec:experiments}

We analyze different evaluation metrics by calculating their score for sentence pairs. 
We use both reference-based metrics, which operate in a monolingual space, and reference-free metrics, which operate in a cross-lingual space. 

\subsection{Metrics}

\paragraph{Reference-based metrics}
We consider the following reference-based metrics. 
\begin{itemize}[topsep=5pt,itemsep=0pt,leftmargin=*]
    \item BERTScore \citep{zhang2019bertscore}
    aggregates and compares BERT embeddings by determining a greedy alignment between words in two sentences and summing up the cosine similarities of representations of aligned words.
    \item MoverScore \citep{zhao2019moverscore} 
    computes an optimal alignment between words in the two sentences using word mover distance \citep{kusner2015word} on top of BERT representations. 
    
    \item Sentence BERT (SBERT) \citep{reimers-2019-sentence-bert} fine-tunes Siamese BERT networks on NLI data, and produces sentence embeddings by using pooling on top of BERT representations. We compute the cosine similarity between SBERT representations. 
    \item SBERT-WK \citep{wang2020sbertwk} 
    is a variant of SBERT which weighs different layers of BERT. 
    \item In contrast to the others, BLEURT \citep{sellam2020bleurt} is a supervised metric and fine-tunes BERT
    on the WMT datasets with available human assessment of translation quality.

\end{itemize}

\paragraph{Reference-free metrics}
We consider the following reference-free metrics: 

\begin{itemize}[topsep=5pt,itemsep=0pt,leftmargin=*]
    
    \item Multilingual Sentence BERT (mSBERT) \citep{reimers-2020-multilingual-sentence-bert} is a multilingual version of Sentence BERT trained on parallel data with multilingual knowledge distillation. A teacher model trained on STS and NLI provides English sentence embeddings. mSBERT (student model) is trained to produce embeddings for the English sentence and its translation which are close to the embeddings of the teacher model.
    
    \item Multilingual Universal Sentence Encoder (MUSE) \citep{yang2019multilingual} is a multilingual sentence embedding. It is a dual-encoder model which was trained on multiple tasks such as NLI and translation ranking. MUSE was trained on mono- and multilingual data. 
    \item LaBSE \citep{feng2020languageagnostic} is a dual-encoder framework. The encoders re-use pre-trained BERT and 
    finetune it with Masked Language Modeling and Translation Language modeling 
    on monolingual and parallel data. 
    \item LASER \citep{artetxe2019massively} is a BiLSTM encoder trained on parallel corpora. It produces language-agnostic representations. The encoder-decoder architecture is trained jointly on different languages.
    \item XMoverScore \citep{zhao-etal-2020-limitations} extends MoverScore to operate in the cross-lingual setup, and relies on re-aligned multilingual BERT representations. Note that we exclude a target-side language model integrated in XMoverScore to have a similar setup as for MoverScore. 
\end{itemize}
Except for XMoverScore, all metrics are based on calculation of cosine similarity between the source-language and target-language sentence embeddings. Except for MUSE and LASER, all metrics are based on BERT representations. 
Note that some multilingual reference-free metrics can also be used in the monolingual reference-based case, especially those based on calculating cosine similarity on top of sentence embeddings, thus we will include them in both settings. 

\subsection{Datasets}
\label{sec:datasets}
We use the datasets of the WMT shared task and the Semantic Text Similarity Benchmark (STSB) in our experiments. In the appendix, Table \ref{tab:dataset_size} shows the statistics in each dataset, and Table \ref{tab:ex_data} shows examples for the sentences of the dataets.

\paragraph{WMT}
The WMT datasets contain an input sentence in the source language, the hypothesis translation of an MT system and a human reference sentence in the target language. 
Humans have rated the similarity between human reference and MT hypothesis using so-called `direct assessment' (DA) scores which are framed in terms of one sentence `adequately expressing the meaning' of another. 
We use these ratings as semantic scores in our setup.

For the reference-based case, we use the hypotheses and the references as sentence pairs. 
This data is collected over multiple language pairs which have English as target language (so both the human reference and the hypothesis are in English) of WMT15-WMT17. 
For the reference-free scenario, we pair the source texts with MT hypotheses and use the corresponding reference-to-hypothesis DA scores for similarity score. 
For German, we take the data from WMT15 \citep{bojar-EtAl:2015:WMT}, WMT16 \citep{bojar-EtAl:2016:WMT1} and WMT17  \citep{bojar-EtAl:2017:WMT1}. Chinese is only available in WMT17.

\paragraph{STSB}
The Semantic Text Similarity Benchmark \citep{cer2017semeval} consists of English sentence pairs and a semantic similarity score for each pair. The scores were annotated by humans. The score is used as semantic score in the regression. In contrast to WMT, some sentence pairs in STSB are designed to have a different structure but a similar meaning.

While we use the sentence pairs directly for  the monolingual case, we translate the sentences in the cross-lingual case using Google Translate, following \citet{chidambaram2018learning}.

\section{Results}

\begin{table*}[!htb]
      \centering
    {\footnotesize
    \begin{tabular}{l|cccc|c}
    \toprule
    \multicolumn{6}{c}{\textbf{WMT}}\\
        \multicolumn{1}{c}{\textbf{Metric}} & \textbf{SEM} & \textbf{SYN} & \textbf{LEX} & \textbf{MOR} & \textbf{$R^2$} \\ 
        \midrule
MoverScore & 0.28 & 0.15 & 0.64 & -0.06 & 0.76\\
BERTScore & 0.27 & 0.16 & 0.61 & -0.01* & 0.74\\
LASER & 0.19 & 0.04 & 0.33 & 0.32 & 0.53\\
SBERT & 0.37 & -0.02* & 0.22 & 0.22 & 0.42\\
SBERT-WK & 0.30 & -0.02 & 0.32 & 0.33 & 0.58\\
LaBSE & 0.31 & 0.00* & 0.36 & 0.22 & 0.53\\
mSBERT & 0.41 & -0.05 & 0.24 & 0.18 & 0.43\\
BLEURT & 0.48 & 0.07 & 0.30 & 0.05 & 0.57\\
mUSE & 0.27 & -0.03& 0.39 & 0.19 & 0.49\\
        \bottomrule
    \end{tabular}
    }
    {\footnotesize
    \begin{tabular}{l|cccc|c}
    \toprule
    \multicolumn{6}{c}{\textbf{STS}}\\
        \textbf{Metric} & \textbf{SEM} & \textbf{SYN} & \textbf{LEX} & \textbf{MOR} & \textbf{$R^2$} \\ 
        \midrule
MoverScore & 0.30 & 0.24 & 0.45 & 0.04 & 0.61\\
BERTScore & 0.12 & 0.11 & 0.67 & 0.06 & 0.69\\
LASER & 0.55 & 0.05 & 0.28 & 0.09 & 0.63\\
SBERT & 0.73 & -0.03 & 0.14 & -0.06 & 0.60\\
SBERT-WK & 0.26 & 0.04 & 0.31 & 0.30 & 0.55\\
LaBSE & 0.60 & 0.07 & 0.27 & 0.06 & 0.67\\
mSBERT & 0.93 & -0.02 & 0.04 & 0.00* & 0.91\\
BLEURT & 0.62 & 0.08 & 0.19 & 0.03* & 0.58\\
mUSE & 0.76 & 0.00* & 0.22 & -0.11 & 0.68\\
        \bottomrule
    \end{tabular}}
    \caption{Regression results for reference-based metrics. Coefficient values for linguistic regressors and $R^2$ values. * denotes $p \geq 0.05$ (non-significance). \wei{Intercept coefficients are small values and we 
    omit them for clarity.}}
    \label{tab:results_mono}
\end{table*}

\paragraph{Reference-based Metrics}
Table \ref{tab:results_mono} shows the results for the reference-based metrics. The $R^2$ values range from 0.43 to 0.76 on WMT, and from 0.58 to 0.91 on STS. 
This means we can reasonably well explain the metrics from our four explanatory variables and using a linear model. mSBERT can best be explained with an $R^2$ value of 0.91 on STS; however, since it has been trained on STS, this merely indicates overfitting. All metrics have positive coefficients for SEM, indicating that they all reflect semantic similarity and (semantic) `adequacy' (as measured by DA), respectively: the weights range from 0.19 to 0.48 on WMT and from 0.12 to 0.76 on STS (ignoring mSBERT). The SYN coefficients are much lower and range from -0.05 to 0.16 for WMT and from -0.03 to 0.24 on STS. MoverScore and BERTScore are most affected by syntactic similarity (0.11 to 0.24), while the sentence embedding based techniques have coefficients around zero. The coefficients for MOR are low on STS, except for SBERT-WK, and moderate for WMT.

\emph{All metrics have comparatively large coefficients for lexical overlap, especially on WMT}: the coefficient values range from 0.24 to 0.64 on WMT and from 0.14 to 0.67 on STS. Especially LEX dominates for MoverScore and BERTScore, indicating that these two metrics are most sensitive to lexical adversaries, potentially making them vulnerable to 
inputs 
such as `man bites dog' vs.\ `dog bites man'.  


\begin{table*}[!htb]
    \centering
    {\footnotesize
    \begin{tabular}{l|ccc|c}
    \toprule
    \multicolumn{5}{c}{\textbf{WMT}}\\
        \textbf{Metric} & \textbf{SEM} & \textbf{SYN} & \textbf{LEX} & \textbf{$R^2$} \\ 
        \midrule
        
XMoverScore & 0.37 & -0.08* & 0.40 & 0.39\\
LASER & 0.29 & -0.03* & 0.38 & 0.30\\
mUSE & 0.34 & 0.01* & 0.35 & 0.33\\
LaBSE & 0.47 & -0.05* & 0.17 & 0.30\\
mSBERT & 0.42 & -0.04* & 0.25 & 0.31\\
        \bottomrule
        \end{tabular}}
    {\footnotesize 
    \begin{tabular}{l|ccc|c}
    \toprule
    \multicolumn{5}{c}{\textbf{STS}}\\
        \textbf{Metric} & \textbf{SEM} & \textbf{SYN} & \textbf{LEX} &  \textbf{$R^2$} \\ 
        \midrule
        
        XMoverScore & 0.09 & -0.12 & 0.46 &  0.24\\
LASER & 0.59 & 0.02* & 0.27 &  0.51\\
mUSE & 0.72 & 0.02* & 0.12 &  0.59\\
LaBSE & 0.63 & 0.08 & 0.26 &  0.56\\
mSBERT & 0.87 & -0.01* & 0.05 &  0.80\\
        
        \bottomrule
        
    \end{tabular}}
    
    \caption{Regression results for Chinese-English reference-free metrics. 
    }
    \label{tab:results_free_zh}
\end{table*}
\begin{table*}[!htb]
    \centering
    {\footnotesize
    \begin{tabular}{l|cccc|c}
    \toprule
    \multicolumn{6}{c}{\textbf{WMT}}\\
        \textbf{Metric} & \textbf{SEM} & \textbf{SYN} & \textbf{LEX} & \textbf{CLB} & \textbf{$R^2$} \\ 
        \midrule
        
XMoverScore & 0.18 & 0.10 & 0.18 & 0.59 & 0.68\\
LASER & 0.18 & -0.02* & 0.32 & 0.34 & 0.43\\
mUSE & 0.26 & 0.00* & 0.25 & 0.30 & 0.40\\
LaBSE & 0.37 & -0.05* & 0.14 & 0.30 & 0.40\\
mSBERT & 0.35 & -0.04* & 0.16 & 0.31 & 0.40\\
        \bottomrule
        \end{tabular}}
    {\footnotesize 
    \begin{tabular}{l|cccc|c}
    \toprule
    \multicolumn{6}{c}{\textbf{STS}}\\
        \textbf{Metric} & \textbf{SEM} & \textbf{SYN} & \textbf{LEX} & \textbf{CLB} & \textbf{$R^2$} \\ 
        \midrule
        
        XMoverScore & 0.08 & 0.16 & 0.36 & 0.48 & 0.53\\
LASER & 0.56 & 0.03* & 0.28 & 0.17 &  0.55\\
mUSE & 0.72 & 0.02* & 0.12 & 0.14 & 0.61\\
LaBSE & 0.63 & 0.09 & 0.25 & 0.10 & 0.58\\
mSBERT & 0.87 & -0.01* & 0.05* & 0.07 & 0.81\\
        
        \bottomrule
        
    \end{tabular}}
    
    \caption{Regression results for Chinese-English reference-free metrics. We add the CLB factor in the regression.
    }
    \label{tab:results_free_zh_clb}
\end{table*}
\paragraph{Reference-free Metrics}

Table \ref{tab:results_free_zh} shows the results for ZH-EN in the reference-free setup (omitting the score for MOR as indicated earlier). Many SYN coefficients are now zero or negative, meaning that a larger syntactic difference between the input arguments leads to a higher metric score, indicating that metrics are sensitive to syntactic language differences. LEX is still significant in all cases. SEM has higher coefficient values than LEX in 6 out of 10 cases, and when it `wins', it wins by a large margin. However, we note that the $R^2$ are low: they range from 0.3-0.39 on WMT and from 0.24-0.59 on STS. This means we can either not (well) explain the metrics given our current regressors or the relationship is not well explained by a linear model. 
The results for DE-EN are similar; 
we provide them in Table~\ref{tab:results_de} (appendix).

To explore why $R^2$ scores are now lower, we note that reference-free metrics based on BERT might contain a form of \textit{cross-lingual bias} (CLB) in that they do not properly score mutual translations, as \citet{Cao:2020} and many others have shown that the multilingual subspaces induced by BERT are mis-aligned.
We thus include 
a factor CLB as a regressor 
to measure how significant this bias is 
in different metrics.
Note that the metrics use either different BERT variants or  other representations such as LASER, which points to different sources of CLB. 
Therefore, we realize CLB differently across metrics. For each metric regression, we use the same metric but take a parallel sentence, i.e., source text and Google translation (as we assume that Google Translate has very high quality in general), as input arguments, and take the metric score as a proxy of the CLB factor.  
If a metric does not contain cross-lingual bias, it should assign almost full scores to parallel sentences; this constant would then be meaningless in the regression.

In Table~\ref{tab:results_free_zh_clb}, 
we show that including the CLB factor in the regression improves the $R^2$. 
We substantially improve the $R^2$ for XMoverScore (from 0.39 to 0.68), but observe little improvements for the remaining metrics (especially on STS). This is because XMoverScore is more problematic than the other metrics in terms of properly scoring mutual translations, given that the other metrics use 
BERT variants (or LASER) that have been finetuned (or trained) on parallel sentences. Apart from CLB, both SEM and LEX are the dominating factors in the regression. The DE-EN results are similar---see Table~\ref{tab:results_de_clb} (appendix).

\paragraph{Limitations}
The $R^2$ scores for the WMT dataset are almost always lower than the corresponding STS scores. 
One may not forget that STS sentences are in a sense artificial sentences of the form `a girl is playing a guitar' while WMT contains more realistic sentences as well as their (possibly faulty, non-grammatical) translations. 
The WMT datasets are furthermore inhomogeneous in that we used different years from 2015 to 2017, which has different participating MT systems as well as slightly different task definitions, corresponding to an aggregation of different domains. The STS dataset is monolingual and was translated by Google Translate for the cross-lingual scenario. 
The latter may lower the quality of the data, but 
WMT data also contains translations. 

The WMT scores measure the similarity between the reference and the hypothesis but  we compare the source with the hypothesis in the cross-lingual scenario, which reflects a mismatch. \citet{fomicheva2020mlqepe} provide a dataset which gives human DA scores between source and hypothesis. We repeated the experiments with this dataset. The full results are shown in Table \ref{tab:res_da} in the appendix \wei{(omitting the CLB factor)}. The $R^2$ scores of 3 out of 5 metrics improve (slightly) compared to the WMT dataset for German-English but all $R^2$ scores are lower for Chinese-English. 
This means that mismatched DA scores are apparently not the main reason for our low regression fits. 
With the new DA scores, all coefficients for SEM 
are considerably lower. 
They are in the range of range 0.06 to 0.14 compared to the maximum of 0.47 for WMT. 
In contrast, all SYN (0-0.12) scores are higher especially vof German. 
All MOR (0.16-0.33) and most LEX scores are higher but they are still in a similar range as for the original DA scores.

\section{Analysis}
\begin{table}[!htb]
\footnotesize
    \centering
    \begin{tabular}{l|ccc}
    \toprule
         & \textbf{Lex(A,B)} & \textbf{Lex(A,C)} & \textbf{Size} \\
         \midrule
         {\small \citet{freitag-bleu-paraphrase-references-2020} 
         }  
         & 0.49 & 0.99 & 1452 \\
         PAWS & 0.84 & 0.94 & 100\\
         \bottomrule
    \end{tabular}
    \caption{Averaged lexical overlap and size of the datasets for the adversarial experiments. 
    }
    \label{table:adv_data}
\end{table}

\begin{table*}[!htb]
    \centering
    \small 
    \begin{tabular}{p{4.5cm}p{4.5cm}p{4.5cm}}
    \toprule
    \multicolumn{3}{c}{\textbf{PAWS}} \\
        \textbf{Sentence A} & \textbf{Sentence B} & \textbf{Sentence C}  \\
        \midrule
        Later in 2014 , Dassault Systèmes was bought by Quintiq. & Dassault Systèmes was bought in 2014 by Quintiq. & In 2014 , Quintiq was bought by Dassault Systèmes.\\
        They are high , built of concrete faced with small blocks of stone. &They are high built of concrete with small stone blocks. & They are small , built of concrete with high stone blocks.\\
        \toprule
        \multicolumn{3}{c}{\textbf{\citet{freitag-bleu-paraphrase-references-2020}} translated} \\
        \textbf{Sentence A} & \textbf{Sentence B} & \textbf{Sentence C}  \\
        \midrule
        Shark injures 13-year-old on lobster dive in California	 & A 13-year-old is injured by a shark while diving for lobsters in California & Shark injures 13-year-old on dive lobster in California\\
        Kovacic did a quick give-and-go at midfield. & Kovacic managed a quick one-two in midfield. & Kovacic did a quick midfield at give-and-go.\\
         \bottomrule
    \end{tabular}
    \caption{Selected sentences for the adversarial experiments.}
    \label{tab:adv_ex}
\end{table*}

\begin{figure*}[!htb]
    \centering
    \includegraphics[width=.75\textwidth]{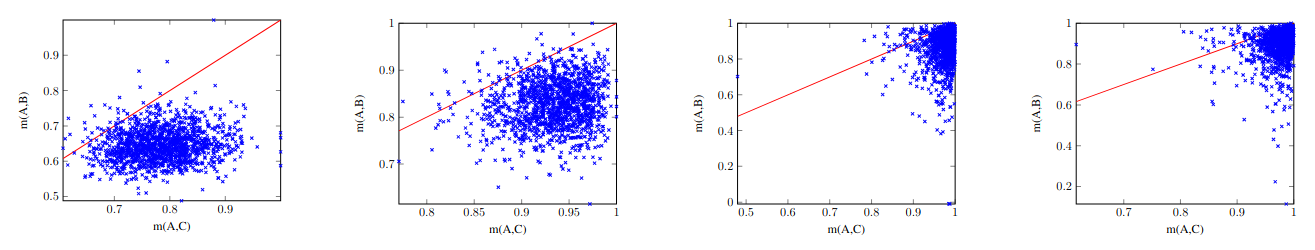}
    \includegraphics[width=.75\textwidth]{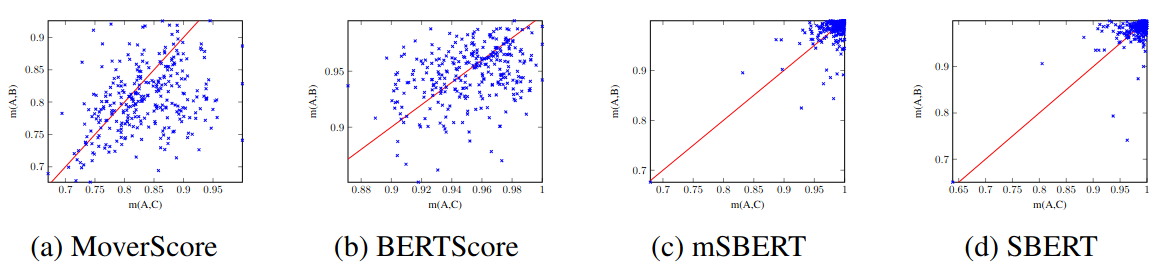}
    \caption{
    Distribution of $\mathfrak{m}(A,B)$ vs.\ $\mathfrak{m}(A,C)$. Top: Dataset from \citet{freitag-bleu-paraphrase-references-2020}. Bottom: PAWS dataset. }
    \label{fig:adv-freitag}
\end{figure*}

In the following, we analyze two observations from our previous experiments in more depth: (i) the sensitivity of metrics to lexical overlap; (ii) the orthogonality of metrics in that they capture different linguistic signals. 

\subsection{Adversarial experiments}
According to our results, all metrics rely on lexical overlap, which indicates that they may not be robust to adversarial examples. 
We check this by an additional experiment, for reference-based metrics, in which we query their pairwise preferences over three sentences: 
Sentence A is the anchor sentence; 
sentence B is a 
paraphrase of sentence A with little lexical overlap;  
sentence C is a non-paraphrase with high lexical overlap. 
A good metric $\mathfrak{m}$ would have $\mathfrak{m}(A,B)  
> \mathfrak{m}(A,C)$ but the high lexical overlap between  A and C makes this task difficult. 

\paragraph{Freitag et al.} Sentences A are taken as  source sentences from WMT19.
\citet{freitag-bleu-paraphrase-references-2020} provided alternative references for WMT19; they instructed human professional translators to paraphrase the references as much as possible in terms of lexical choice and sentence structure but keep the same semantics. We take these as sentences B.  
We produce sentences C from sentences A: 
for each sentence A, we detect the nouns within the sentence using the NLTK POS tagger, and then we randomly permute them to produce a sentence C.
Since \citet{freitag-bleu-paraphrase-references-2020} provided sentences in German, we translate all sentences into English using Google translate. We note that by inspection the translations are generally of high quality and satisfy our constraints of inducing paraphrases with low lexical overlap and non-paraphrases with high lexical overlap. 
Examples are shown in Table \ref{tab:adv_ex} and statistics in Table \ref{table:adv_data}.

\paragraph{PAWS} We complement the analysis with the native English PAWS dataset \citep{paws2019naacl}
which consists of 
paraphrase and non-paraphrase pairs that have high lexical overlap. 
For each sentence in the dataset, there 
are multiple paraphrases and non-paraphrases. 
For a given sentence A, we use the paraphrase with the smallest amount of lexical overlap as sentence B 
and sentence C is the non-paraphrase with the highest amount of lexical overlap with A. 
We note that PAWS is more problematic as even the paraphrases B with minimum amount of lexical overlap do have considerable lexical overlap. Therefore, we select the 100 sentence pairs with the smallest amount of lexical overlap between sentence A and B. Table \ref{table:adv_data} shows the lexical overlap and the size of the datasets. 
Indeed, for PAWS,
sentences B have 
only a little less lexical overlap with A than sentences C,  
while 
the dataset of \citet{freitag-bleu-paraphrase-references-2020} 
has a much clearer separation between B and C in this respect.

Figure \ref{fig:adv-freitag} shows the distribution of $\mathfrak{m}(A,B)$ and $\mathfrak{m}(A,C)$ for selected metrics $\mathfrak{m}$. 
Overall, the adversarial results on translated and non-translated datasets point in a similar direction.
We see that metrics clearly prefer the high lexical overlap sentences which are non-paraphrases (sentences C) in the translated dataset of \citet{freitag-bleu-paraphrase-references-2020}. For \wei{non-translated} PAWS, metrics 
are at least to some degree indifferent, but tend to prefer B on average, with MoverScore and BERTScore having higher preference for C than mSBERT and SBERT, which confirms our linear regression results. 

Overall, these experiment show that metrics are indeed not robust to lexical adversarial examples. 

\subsection{Ensemble of Models}
Our experiments in Section \ref{sec:experiments} show that the different metrics 
use different information signals, even when they use the same underlying BERT representations. 
For example, BERTScore relies more on lexical overlap and mSBERT relies more on semantics; BERTScore and MoverScore both capture syntax, while the other metrics are less sensitive to it. This means that the metrics are to some degree orthogonal.  
Thus, we suspect that a combination of metrics yields a considerably better metric. 
We check this hypothesis through an extra experiment. 
We combine especially BERTScore and MoverScore with SBERT and mSBERT based metrics. 

We evaluate the metrics on segment-level.
In segment-level evaluation, each sentence pair gets a score from $\mathfrak{m}$. 
The Pearson correlation coefficient is then calculated between 
these scores and the human judgement, disregarding the systems which generated the translations.
To combine metrics, we simply average their scores. 
In the evaluation, we use the best performance of the single metrics as baseline and compare it to our combined metrics. 

Table \ref{tab:ensemble} (appendix) shows the improvements for different language pairs. 
In the reference-free case, 
the two best ensembles combine XMoverScore with mSBERT and LaBSE---the latter two rely less on lexical overlap than the first---leading to big improvements of 11-13\% over the best individual metric. 
Combining metrics relying on similar factors shows less improvement, and often even leads to worse results. In the reference-based case, we combine BERTScore with mSBERT and observe an improvement of 8\%, more than for any other combination we tested.  
These results show that combing metrics relying on different factors can largely improve their performance.\footnote{In work independent from us, \citet{song-etal-2021-sentsim} also ensemble BERT based reference-free evaluation metrics.}

\section{Conclusions}

We disentangled BERT-based evaluation metrics along four linguistic factors: semantics, syntax, morphology, and lexical overlap. The results indicate that (i) the different metrics capture these different aspects to different degrees but (ii) they all rely on semantics and lexical overlap.  The first observation indicates that combining metrics may be helpful, which we confirmed: simple parameter-free averaging of hetereogenous metric scores can improve correlations with humans by up to more than 13\% in our experiments. The second  observation shows that these metrics may be prone to adversarial fooling, just like BLEU and ROUGE, which we confirmed in an additional experiment in which we queried metric preferences over paraphrases with little lexical overlap and non-paraphrases with high lexical overlap. 
Future metrics should especially take this last aspect into account, and improve their robustness to adversarial conditions. 

There is much scope for future research, e.g., in developing better global explanations for reference-free metrics (as we cannot yet well explain these metrics), better linguistic factors (e.g., a clearer conceptualization of morphological similarity of two sentences) and in developing local explainability techniques for evaluation metrics \citep{fomicheva:etal:2021:eval4nlp,fomicheva2021translation}.\footnote{Code and data for our experiments are available from \url{https://github.com/SteffenEger/global-explainability-metrics}.}

\section*{Acknowledgments}
We thank the anonymous reviewers for their insightful comments and suggestions, which greatly improved the final version of the paper. We also thank Micha Dippell for his early experiments contributing to this work. 
This work has been supported by the German Research Foundation as part of the Research Training
Group Adaptive Preparation of Information from Heterogeneous Sources (AIPHES) at the Technische
Universit\"at Darmstadt under grant No. GRK 1994/1.

\bibliography{emnlp2021}
\bibliographystyle{emnlp_natbib}

\clearpage
 \section{Appendix}
 The following tables contain remaining experimental results.
\begin{table}[!htb]
    \centering
    \footnotesize
    \begin{tabular}{c|ccc}
    \toprule
        \textbf{Dataset} & \textbf{English} & \textbf{Chinese} & \textbf{German} \\ \midrule
        WMT & 8595 & 560 & 1620\\
        STSB & 5749 & 5749 & 5749\\
        \bottomrule
    \end{tabular}
    \caption{Number of sentence pairs in each dataset and language pair.}
    \label{tab:dataset_size}
\end{table}

\setlength{\tabcolsep}{3.5pt}
\begin{table}[!htb]
    \centering
    \footnotesize
    {
    \begin{tabular}{llc}
    \toprule
    \multicolumn{3}{c}{\textbf{WMT}}\\
        \textbf{Sentence 1} & \textbf{Sentence 2} & \textbf{SEM}  \\
        \midrule
         Why is it such a difference? & Why such a difference? & 0.94\\
         There are coincidences. & Zufälle gibt es. & 0.57\\
         \bottomrule\\
         \multicolumn{3}{c}{\textbf{STS}}\\
         \midrule
         Some men are fighting.	& Two men are fighting. & 4.25\\
         A woman is writing. & Eine Frau schwimmt. & 0.1\\
         
         \bottomrule
    \end{tabular}
    }
    \caption{Example of WMT and STS datasets.}
    \label{tab:ex_data}
\end{table}

\begin{table*}
      \centering
      {\footnotesize
    \begin{tabular}{l|cccc|c}
    \toprule
    \multicolumn{6}{c}{\textbf{WMT}}\\
        \textbf{Metric} & \textbf{SEM} & \textbf{SYN} & \textbf{LEX} & \textbf{MOR} & \textbf{$R^2$} \\ 
        \midrule
        
XMoverScore & 0.37 & -0.07* & 0.33 & -0.01* & 0.31\\
LASER & 0.21 & -0.08 & 0.31 & 0.31 & 0.29\\
mUSE & 0.19 & -0.00* & 0.40 & 0.08 & 0.25\\
LaBSE & 0.30 & -0.03* & 0.26 & 0.07 & 0.21\\
mSBERT & 0.39 & -0.06* & 0.22 & 0.05* & 0.26\\
        \bottomrule
        \end{tabular}}\hspace{0.3cm}
    {\footnotesize
    \begin{tabular}{l|cccc|c}    
        \toprule
    \multicolumn{6}{c}{\textbf{STS}}\\
        \textbf{Metric} & \textbf{SEM} & \textbf{SYN} & \textbf{LEX} & \textbf{MOR} & \textbf{$R^2$} \\ 
        \midrule
        
        XMoverScore & 0.08 & -0.02* & 0.51 & 0.03 & 0.30\\
LASER & 0.54 & 0.00* & 0.28 & 0.15 & 0.54\\
mUSE & 0.71 & 0.00* & 0.09 & 0.07 & 0.6\\
LaBSE & 0.60 & 0.05 & 0.29 & 0.04 & 0.60\\
mSBERT & 0.89 & -0.02 & 0.05 & 0.02 & 0.84\\
        
        \bottomrule
        
    \end{tabular}}
    \caption{Regression results for German-English reference-free metrics. Coefficient values for linguistic regressors and $R^2$ values. * denotes $p \geq 0.05$ (non-significance).}
    \label{tab:results_de}
    
\end{table*}

\begin{table*}[!htb]
    \centering
    {\footnotesize
    \begin{tabular}{l|ccccc|c}
    \toprule
    \multicolumn{7}{c}{\textbf{WMT}}\\
        \textbf{Metric} & \textbf{SEM} & \textbf{SYN} & \textbf{LEX} & \textbf{MOR} & \textbf{CLB} & \textbf{$R^2$} \\ 
        \midrule
        
XMoverScore & 0.14 & 0.08 & 0.16 & -0.03* & 0.63 & 0.61\\
LASER & 0.11 & -0.03* & 0.28 & 0.19 & 0.40 & 0.46\\
mUSE & 0.19 & 0.02* & 0.28 & 0.03* & 0.35 & 0.34\\
LaBSE & 0.29 & 0.00* & 0.14 & 0.00* & 0.45 & 0.38\\
mSBERT & 0.36 & -0.02* & 0.09 & 0.02* & 0.36 & 0.36\\
        \bottomrule
        \end{tabular}}\hspace{0.1cm}
    {\footnotesize 
    \begin{tabular}{l|ccccc|c}
    \toprule
    \multicolumn{7}{c}{\textbf{STS}}\\
        \textbf{Metric} & \textbf{SEM} & \textbf{SYN} & \textbf{LEX} & \textbf{MOR} & \textbf{CLB} & \textbf{$R^2$} \\ 
        \midrule
        
        XMoverScore & 0.09 & 0.25 & 0.43 & -0.09 & 0.40 & 0.54\\
LASER & 0.51 & 0.01* & 0.32 & 0.12 &  0.11 & 0.58\\
mUSE & 0.70 & 0.02* & 0.12 & 0.05* & 0.11 & 0.61\\
LaBSE & 0.59 & 0.06 & 0.30 & 0.02* & 0.12 & 0.62\\
mSBERT & 0.88 & -0.02* & 0.05* & 0.02* & 0.03* & 0.85\\
        
        \bottomrule
        
    \end{tabular}}
    
    \caption{Regression results for German-English reference-free metrics. Coefficient values for linguistic regressors and $R^2$ values. * denotes $p \geq 0.05$ (non-significance). We add the CLB factor in the regression.}
    \label{tab:results_de_clb}
\end{table*}


        

        
        
        
    
\begin{table*}[!htb]
    \centering
    {\footnotesize
    \begin{tabular}{l|cccc|c}
    \toprule
    \multicolumn{6}{c}{\textbf{German-English}}\\
        \textbf{Metric} & \textbf{SEM} & \textbf{SYN} & \textbf{LEX} & \textbf{MOR} & \textbf{$R^2$} \\ 
        \midrule
        
        XMoverScore & 0.16 & 0.05* & 0.33 & 0.16 & 0.20\\
LASER & 0.12 & 0.08 & 0.30 & 0.46 & 0.41\\
mUSE & 0.12 & 0.12 & 0.25 & 0.43 & 0.35\\
LaBSE & 0.10 & 0.08 & 0.23 & 0.51 & 0.41\\
mSBERT & 0.14 & 0.06* & 0.21 & 0.33 & 0.22\\
        \bottomrule
    \end{tabular}}\hspace{0.3cm}
    {\footnotesize
    \begin{tabular}{l|cccc|c}
        \toprule
    \multicolumn{6}{c}{\textbf{Chinese-English}}\\
        \textbf{Metric} & \textbf{SEM} & \textbf{SYN} & \textbf{LEX} &  & \textbf{$R^2$} \\ 
        \midrule
        
        XMoverScore & 0.14 & 0.04* & 0.35 &  & 0.15\\
LASER & 0.06* & 0.03* & 0.35 &  & 0.14\\
mUSE & 0.12 & 0.05* & 0.42 &  & 0.21\\
LaBSE & 0.16 & 0.04* & 0.43 &  & 0.23\\
mSBERT & 0.24 & 0.00* & 0.31 &  & 0.17\\

        \bottomrule
    \end{tabular}}
    \caption{Regression results for reference-free metrics on the  \citet{fomicheva2020mlqepe} dataset which contains DA scores comparing sources and hypotheses (rather than references and hypotheses). Coefficient values for linguistic regressors and $R^2$ values. * denotes $p \geq 0.05$ (non-significance).}
    \label{tab:res_da}
\end{table*}
\begin{table*}[!htb]
\footnotesize
    \centering
    \begin{tabular}{l|cccccccc}
    \toprule
    & \textbf{cs-en} & \textbf{de-en} & \textbf{fi-en} & \textbf{lv-en} & \textbf{ru-en} & \textbf{tr-en} & \textbf{zh-en} & \textbf{Avg}\\
    \midrule
    \textbf{Reference-based}\\
    BERTScore + mSBERT & 8\% & 15\% & 7\% & 7\% & 11\% & 2\% & 4\% &  8\% \\
    SBERT + MoverScore & 3\% & 10\% &1\% &11\% &6\% & -1\% & 1\% & 4\% \\
    SBERT + LASER & 6\% & 6\% & 0\% & 2\% & 5\% & 3\% & 10\% & 5\% \\ 
    SBERT + mUSE & 5\% & 10\% & 4\% & 2\% & 8\% & 4\% & 2\% & 5\% \\
     \textbf{Reference-free}\\
     XMoverScore + mSBERT & 9\% & 19\% & 11\% & 12\% & 18\% & 13\% & 12\% & 13\% \\
     XMoverScore + LaBSE & 8\% & 14\% & 5\% & 10\% & 17\% & 9\% & 17\% & 11\% \\
     mSBERT + LASER & 8\% & 9\% & -2\% & 1\% & 3\% & 0\% & 12\% & 4\% \\
    LaBSE + LASER & 7\% &15\% & -4\% & 2\% & 1\% & -2\% & 4\% & 3\% \\
XMoverScore + LASER & 0\% & 7\% & -5\% & 2\% & 4\% & 1\% & 13\% & 3\% \\
mSBERT + LaBSE & 3\% & 7\% & -11\% & -6\% & -7\% & -6\% & 2\% & -3\% \\
    \bottomrule
      
    \end{tabular}
    \caption{Performance gains from the ensemble metrics over single best metrics.}
    \label{tab:ensemble}
\end{table*}

\end{document}